\documentclass[letterpaper, 10 pt, conference]{ieeeconf}  

\IEEEoverridecommandlockouts                              
\usepackage{amsmath} 
\usepackage{amssymb}  
\usepackage{verbatim}
\usepackage{todonotes}
\usepackage{enumerate,amsmath}
\usepackage{graphicx}
\usepackage{color}
\usepackage{cite}
\title{\LARGE \bf
Information Processing Capability of Soft Continuum Arms
}

\author{Estefany A. Torres$^{1}$, Kohei Nakajima$^{2, 3}$, and Isuru S. Godage$^{1}$
\thanks{$^{1}$School of Computing, DePaul University, Chicago, IL 60604, USA.
        {\tt\small igodage@depaul.edu}}%
\thanks{$^{2}$Graduate School of Information Science and Technology, The University of Tokyo, 7-3-1 Hongo, Bunkyo-ku, 113-8656 Tokyo, Japan
        }%
\thanks{$^{3}$JST, PRESTO, Kawaguchi, Saitama 332-0012, Japan
        {\tt\small k\_nakajima@mech.t.u-tokyo.ac.jp}}%
\thanks{This work was supported in part by the National Science Foundation (NSF) grant number~1718755, and by JST PRESTO Grant Number JPMJPR15E7, Japan, and JSPS KAKENHI Grant Numbers JP18H05472, JP16KT0019, and JP15K16076.}
}

\begin{document}

\maketitle
\thispagestyle{empty}
\pagestyle{empty}

\begin{abstract}
	Soft Continuum arms, such as trunk and tentacle robots, can be considered as the "dual" of traditional rigid-bodied robots in terms of manipulability, degrees of freedom, and compliance. Introduced two decades ago, continuum arms have not yet realized their full potential, and largely remain as laboratory curiosities. The reasons for this lag rest upon their inherent physical features such as high compliance which contribute to their complex control problems that no research has yet managed to surmount. Recently, reservoir computing has been suggested as a way to employ the body dynamics as a computational resource toward implementing compliant body control. In this paper, as a first step, we investigate the information processing capability of soft continuum arms. We apply input signals of varying amplitude and bandwidth to a soft continuum arm and generate the dynamic response for a large number of trials. These data is aggregated and used to train the readout weights to implement a reservoir computing scheme. Results demonstrate that the information processing capability varies across input signal bandwidth and amplitude. These preliminary results demonstrate that soft continuum arms have optimal bandwidth and amplitude where one can 
	implement reservoir computing.
\end{abstract}

\section{INTRODUCTION}\label{sec:intro}

The advancement of bio-inspired soft continuum robots, featuring high compliance and inherent safety of operation in contrast to traditional rigid-bodied, precise but often dangerous robots, opens up novel research paradigms \cite{Soft_Nature2015,Soft_ScienceRobo2016,Soft_SoRo2018}. Continuum robotics is an umbrella term that herein is used to cover all types of active and physically reactive compliant systems. In this paper, we mainly focus on continuum robotic systems that can be bent, twisted and elongated, actively or passively, during operation. Continuum robots in this sense have often been made of elastomeric material or superelastic alloys, allowing them to change their shape with a few degrees of freedom (DoF) to form complex “organic” shapes (in contrast to fixed “geometric shapes” of rigid-bodied robots) and to be theoretically able to regulate stiffness over a broad range. Because of the passive deformation these robots undergo in the face of external forces, they can be considered as infinite DoF systems that are highly under-actuated. Continuum robotics has been a highly active area of research in the past few years. An impressive number of prototypes has been proposed over the years \cite{mcmahan2006field,grzesiak2011bionic,webster2010design,mcmahan2005design}. Yet control of continuum robots has not received as much attention, and therefore much of the control of continuum robots is limited to open loop control. Due to the presence of infinitely many DoF associated with inherent compliance, it became clear that controlling continuum robots are challenging. Kinematic control has been implemented on continuum robots wherein the robots are moved slowly in order not to invoke the compliance related oscillatory behaviors \cite{mahl2014variable}. As a result, despite their enormous potential as human-friendly manipulators, continuum arms are primarily restricted to lab spaces, with their 
full potential 
remained to be realized.

\begin{figure}[t]
	\begin{centering}
		\includegraphics[width=1\columnwidth]{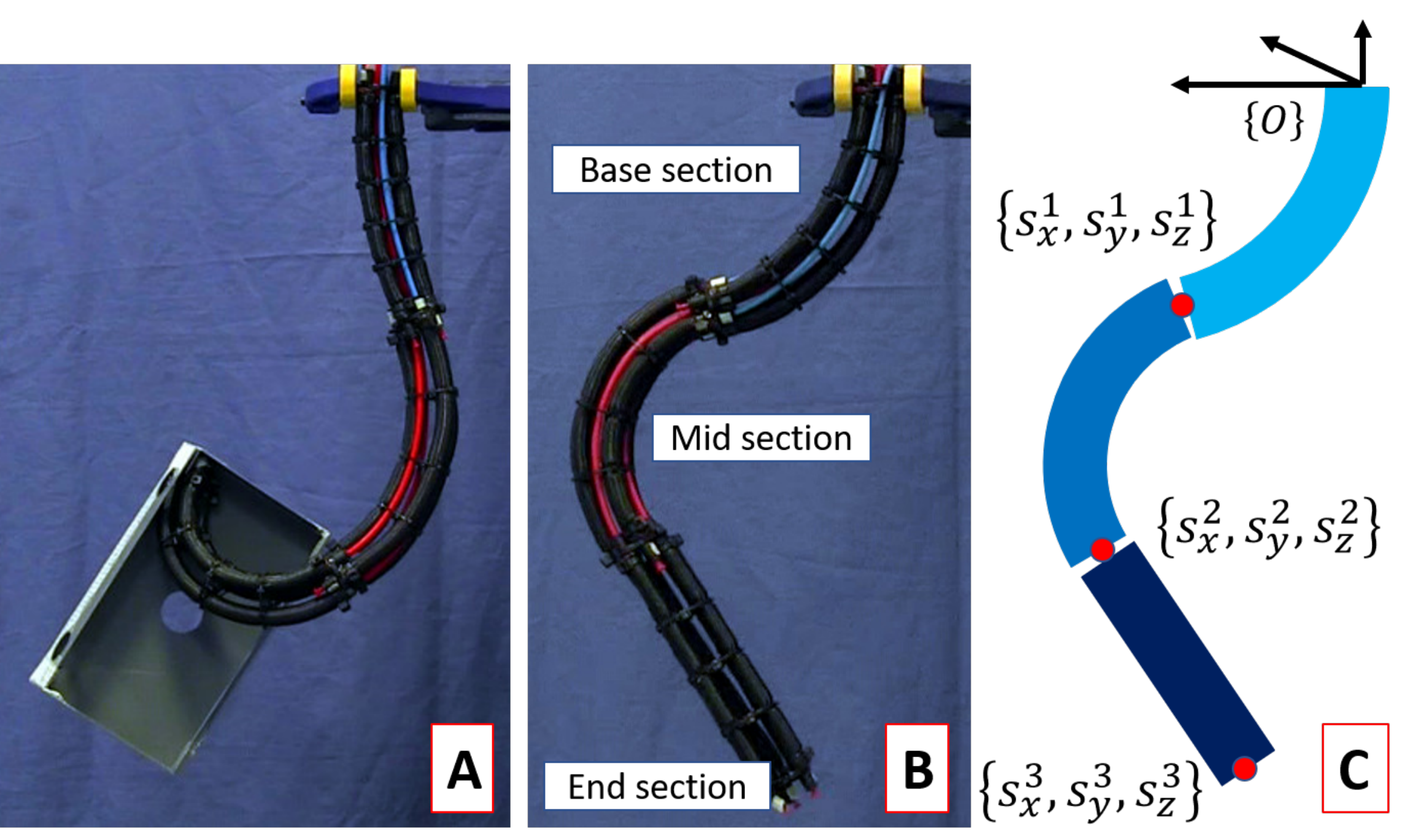}
		\par\end{centering}
	\caption{(A) Pneumatically actuated multisection continuum arm designed at Italian Institute of Technology handling and object, (B) The same arm bending on a plane \cite{godage2015modal}, and (C) Schematic of the prototype arm showing the coordinate convention and the output signal points (tip coordinates) used to record system response in this paper.}
	\label{fig:scenario}
\end{figure}

Related to complex and compliant system control, a potentially ground breaking application is the study of morphological
computations in which a robotic system uses its body dynamics (that
are dependent on the configuration and physical properties such as
stiffness and damping) to simplify control computations \cite{pfeifer2006morphological,Helmut_MC2011}.
In other words, some aspects of control can be outsourced to the body,
using it as a computational resource, with the  notion that these dynamic functions are already
\textquotedblleft encoded\textquotedblright{} within it. This approach
can drastically reduce the complexity of the robot\textquoteright s
system dynamics computation and the corresponding control problems. A classic example
of such embodied morphological intelligence related locomotion control
is observed when the body of a dead fish starts swimming upstream
when exposed to flowing water \cite{beal2006passive}. We can conclude
that a fish can \textquotedblleft outsource\textquotedblright{} the
bulk of the neuromuscular coordination tasks to the \textquotedblleft body\textquotedblright{}, without continuous oversight by the central nervous system, 
when the environmental conditions (hydrodynamic vortices) are right. 

Recent works on morphological computational aspects for soft bodies \cite{Kohei_ICRA2013,Kohei_soft2013,Qian_IROS2013,Kohei_soft2014,Ken_Interface2014,Kohei_soft2015b,Kohei_soft2018,Kohei_book_soft2019} has shown that the mechanical properties of such infinite-dimensional bodies can be exploited as real-time computational assets.
This approach is related to the reservoir computing framework \cite{Reservoir_2002,Reservoir_2004,Reservoir_2007}. 
The idea of a reservoir computing is to map inputs into a high-dimensional space so that the features of input signal can be linearly separable. 
This is achieved by training a learning system solely on readout signals and as a result it is ideally fit to understand the relationship of complex high dimensional physical systems such as soft continuum arms. Once a successful mapping is identified, the physical system itself can be used to compute and predict the behavior of the physical system, which could be used for controlling without the need of external controller, such as recurrent neural networks. 
The emulatability (or expression power) of such mapping is termed as the systems' information processing capability in this paper.
Another key benefit of reservoir computing is that it is faster to train than traditional recurrent neural networks due to the simple learning schemes associated with it. %
In this paper, we will apply reservoir computing principles to investigate the information processing capability of the soft continuum arm shown in Fig.~\ref{fig:scenario}-A~\&~B.

\section{SYSTEM MODEL}\label{sec:model}

\subsection{Prototype Description}\label{sub:proto}

Each of the three continuum sections of the arm shown in Fig.~\ref{fig:scenario}-A
are powered by three pneumatic muscle actuators (PMA). Each PMA is
0.15~m when unactuated and can extend up to 0.065~m at 600~kPa
input pressure. A PMA is constructed from silicone rubber tube (inner
diameter 7.5~mm and outer diameter 9.5~mm) as the PMA containment
layer and polyester braided sheath (7~mm to 17~mm diameter range)
as the outer layer to control the radial expansion and obtain extension.
Nylon union tube connectors of inner diameter 4~mm are used to seal
the Silicone tubes and facilitate air flow. More information about
the fabrication of PMAs can be found in \cite{godage2012pneumatic}.
The %
PMAs of a single continuum section are mounted at 120$^{\circ}$ apart
from each other and constrained in such a way that they maintain 0.0125~m
length from the neutral axis (a hypothetical line that runs in the
center of the arm cross-section). The 3D printed joints that connect
adjacent continuum sections introduce a 60$^{\circ}$ angle offset
about the neutral axis ($+Z$ axis of $\left\{ O_{i}\right\} $).
We use rigid plastic constrainers to maintain the PMAs in parallel to
the neutral axis and provide improved torsional stiffness to minimize
the possible deviations from constant-curvature deformation when operating 
under the influence of gravity. Each continuum section,
inclusive of the tubing and constrainers has an approximate mass of
$0.13$$kg$. %

The pressure supply to the PMAs are precisely controlled by 9 digital
proportional pressure regulators at 20~Hz that output pressures proportional to analog voltage signals between 0-10~V. The voltage signals are
generated using a National Instrument PCI-6704 data acquisition card.
The continuum arm motion in the workspace is tracked by using a Polhemus
G4 wireless magnetic tracking system. The tracking system allows high speed
6~DoF motion tracking at 100~Hz without subjected to occlusions (due
to complex deformations) common in image-based tracking. In total,
we mounted four trackers (two on the joints and one each on the origin
and the tip of the end-section, see Fig. \ref{fig:scenario}-C). The sensing and actuation are controlled
by a Matlab Desktop Realtime Simulink model implemented on Matlab
2018a.

\subsection{Dynamic Model}\label{sub:dyna_model}

The information capacity search requires large number of data, typically
involving more than 5000 randomly generated sample points at
different amplitudes and fundamental signal frequencies (bandwidths). As reported
in \cite{godage2012pneumatic,godage2016dynamics} PMAs have a pressure
deadzone, close to 100~kPa, where little to no extension is observed.
The reason is that, in the pressure deadzone, the Silicon bladders
undergo radial expansion within the braided mesh. Thus, the pressure
range mapped to length changes is {[}100~kPa,600~kPa{]} and we define
6 pressure amplitudes in that range (in 100~kPa intervals). Similarly,
given the prototype arm's oscillatory behavior and the ability to respond quickly
to fast input pressure signals, we define 7 time periods (1/8, 1/4,
1/2, 1, 2, 3, and 4) in which we will generate the systems responses.
Further, for each combination of amplitude and time period, it is recommended to
conduct number of trials to remove any bias from the input signals. Consequently,
generating these results on the prototype robotic arm, particularly
in the proof of concept stage, is not feasible. For instance, to generate
20 trials of 5,000 samples at fundamental period (bandwidth) 4~s, it will take $4\times10^{5}$~s.
Therefore, the dynamic model developed for the prototype arm, detailed
in \cite{godage2016dynamics}, is used to generate the dynamic responses.
The dynamic model was developed using the principles of integral Lagrangian
formulation and runs at real-time efficiency. The model has been experimentally
validated for step pressure responses, which included out-of-plane
bending. In addition, the model, which was based on the constant-curvature assumption,
maintained that assumption throughout those experiments. The model was implemented on
Matlab Simulink 2018a. The input signals were then randomly generated,
applied to the Simulink model, and the joint positions were calculated
(joint-space variables are applied to the kinematic model \cite{godage2015modal}),
and recorded. These data are then used in the processing stage, detailed
in Sec.~\ref{sub:calc_IP}.

\section{METHODOLOGY}\label{sec:method}

\subsection{Dynamic Response Generation}\label{sub:responseGeneration}


The dynamic model of the prototype soft continuum arm shown in Fig.~\ref{fig:scenario}, as detailed in \cite{godage2016dynamics}, is given
by

\begin{align}
	\mathbf{M}\boldsymbol{\ddot{q}}+\mathbf{C}\dot{\boldsymbol{q}}+\mathbf{D}\dot{\boldsymbol{q}}+\boldsymbol{G} & =\boldsymbol{F}\label{eq:dynaModel}
\end{align}
where $\mathbf{M}\in\mathbb{R}^{9\times9}$ is the generalized inertia
matrix, $\mathbf{C}\in\mathbb{R}^{9\times9}$ is the centrifugal/Coriolis
force matrix, $\boldsymbol{G}\in\mathbb{R}^{9}$ is the conservative
force vector, and $\boldsymbol{F}\in\mathbb{R}^{9}$ is the input
force vector. Physically, $\boldsymbol{F}$ is the forces generated
by the fluidic actuators.

We first generate a random input signal $u$ with time steps of $\tau$.
Any $u\left(k\right)$ value of the input signal is kept constant
until the next sample, $u\left(k+1\right)$. 

\begin{figure}[t]
	\begin{centering}
		\includegraphics[width=85mm]{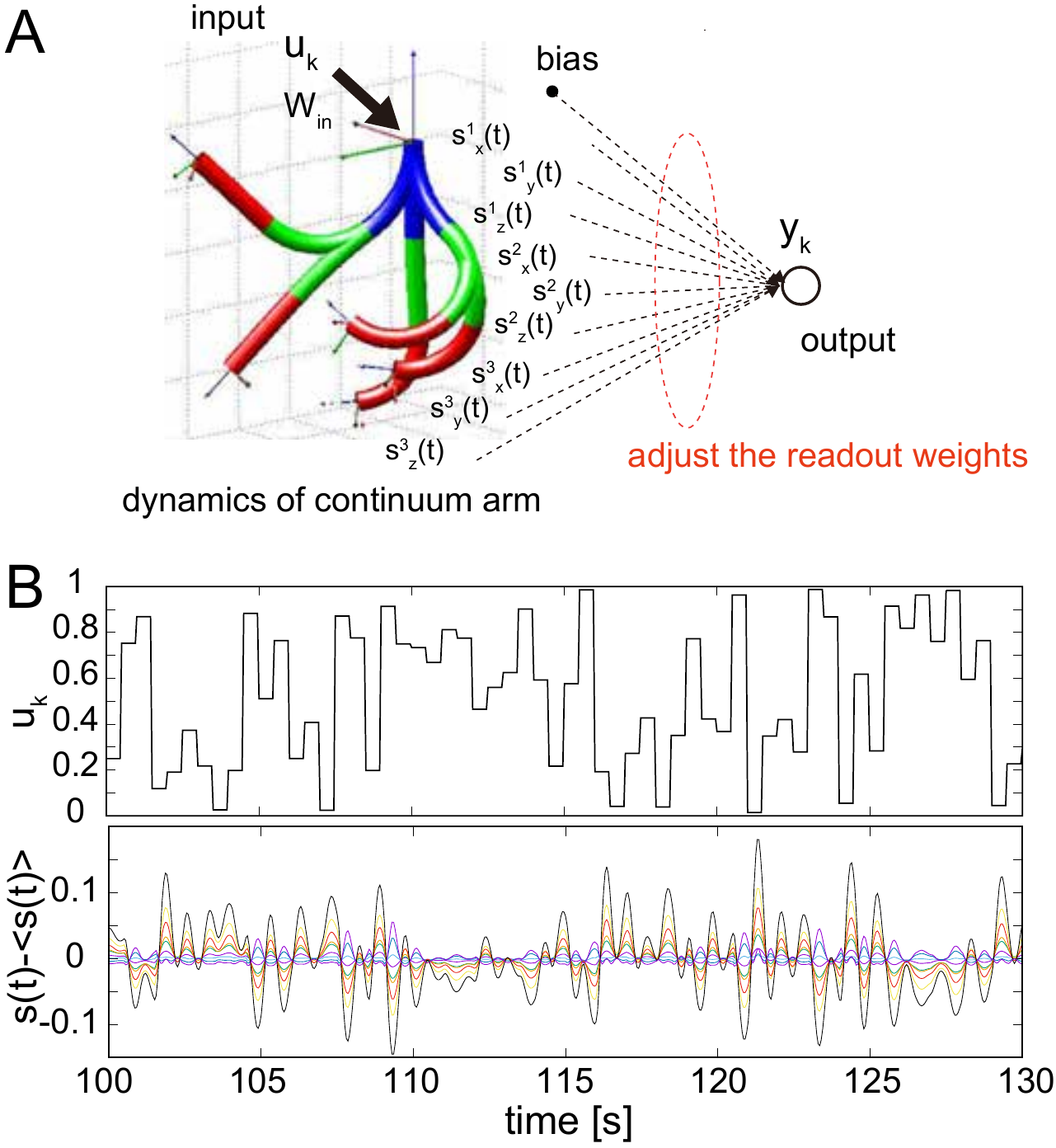}
		\par\end{centering}
	\caption{Schematics showing the information processing scheme using the arm (A) and the typical body dynamics generated by the random motor commands (B). For (B), the upper plot shows the input series and the lower plot shows the corresponding normalized sensory time series. The parameter of the arm is set to $(A, \tau)$ = (2, 0.5).}
	\label{fig:dynamic_model_schematics}
\end{figure}

Upon applying the input
signal (pressures) to relevant PMAs, we measure the system response
(i.e., motion) of the arm. We measure the position coordinates of
each of the continuum section tips (see Fig. \ref{fig:scenario}-C) and recorded at
$\tau/10$ bandwidth. The choice for $10\times$ sampling rate (relative
to the input signal bandwidth) ensures that the readout data is faithful to
the actual motion of the arm in the range of $\tau$ we consider
in this work. Otherwise, due to the fast dynamics of the prototype
continuum arm (as documented in \cite{godage2016dynamics}), the recorded
data will be incomplete and yields incorrect results. For instance, Fig. \ref{fig:dynamic_model_schematics}-B shows the plots of the input (top) and output coordinate (bottom) trajectories for $\tau=0.5$. It can be seen that the smooth output data will be lost had the data were recorded at the input signal bandwidth $\tau=0.5$ instead at $\tau=0.05$.

\subsection{Information Processing using the soft continuum arm}\label{sub:calc_IP}
We here explain how to use our soft continuum arm as a reservoir. 
In this work, we only consider the base section of the continuum arm
(see Fig. \ref{fig:scenario} The reason for this selection is two fold. First,
it reduces the complexity of the input signal (3 in contrast to 9
signals). Second, because the sections are attached serially, base
section has influence over the successive sections. As a result, we
can observe how the physical excitation propagates along the length
of  arm via the compliant structure. 

The input stream is injected to our system as addition of forces to the PMA actuators (also can be considered as active springs due to their high compliance) of the base section (the nearest to the base) with input weights $\bold{W_{in}} = [w_{x} \  w_{y} \  w_{z}]^{T}$.
Throughout our analysis in this study, we use random real value in the range of [0, 1] for the input $u_{k}$ at timestep $k$.
This is to avoid adding temporal correlations from the external signals, which is important when analyzing the information processing capability of figures/the arm itself.   
Furthermore, to see the effect of the amplitude of the input forces, input weights are assigned randomly from the range of $0<w_{x}, w_{y}, w_{z} < A$, and the parameter $A$ is controlled. This input is then transformed
to a pressure signal by multiplying the amplitude of the pressure
signal, $A$. The forces acting on the PMAs are then found by calculating
the area over which the pressure is applied, $A\pi r^{2}$ where $r=8\,mm$
is the mean radius of PMAs during actuation.
The corresponding outputs to this input stream are generated by the weighted sum of the joint coordinates (we call these, sensory values, $\{ s_{x}^{i}, s_{y}^{i}, s_{z}^{i} \}$, for section $i$), which act as computational nodes in our setup (Fig. \ref{fig:dynamic_model_schematics}A).
This implies that we have three sensory values for each section, and since we have three sections in this study, we have nine sensory values in total (Fig. \ref{fig:dynamic_model_schematics}B).

Here, we set parameter $\tau$ to regulate the timescale of each I/O computation (that is, a single timestep), in which a single computation takes the time range of $\tau$ [s] (Fig. \ref{fig:dynamic_model_schematics}B).
Accordingly, we divide each time range $\tau$ into 10 fragments and correspond each sensory value (e.g., $\{ s_{x}^{1}(k\tau +(\tau/10)), s_{x}^{1}(k\tau +2*(\tau/10)),...,    s_{x}^{1}(k\tau + 10*(\tau/10) \}$) to the input $u_{k}$.
This implies that 10 values are fragmented for each sensory value, which samples $9 \times 10 =90$ values in total.
Using these sampled data, 90 computational nodes $\{ x_{k}^{1}, x_{k}^{2}, ..., x_{k}^{i}, ..., x_{k}^{90} \}$ in total were prepared with reconfigured numbering $i$.
Now, according to the inputs $u_{k}$ provided to the system, the corresponding output $y_{k}$ is calculated as $y_{k} = \sum_{i=0}^{90} w_{out}^{i}x_{k}^{i}$, where $x_{k}^{0}$ is set to 1 as a bias term and $w_{out}^{i}$ is the readout weight of the $i$-th computational node.

In the reservoir computing framework, the learning of the target function $\hat{y}_{k}$ is conducted by adjusting the linear readout weights $w_{out}^{i}$. 
During the training phase of the weights, the input stream is provided to the system, which then generates the arm motions, and the corresponding sensory time series is collected together with the target outputs for supervised learning. 
In this study, we apply a ridge regression, which is known as an L2 regularization, to obtain the optimal weights.
In the evaluation phase, using the optimal readout weights, you drive the system with a new input stream and generate the system output, which is then used for the analysis of the performance. 
Throughout our experiments in this study, the washout phase, the training phase, and the evaluation phase are set to 500 timesteps, 2000 timesteps, and 2500 timesteps, respectively. 

\section{RESULTS}\label{sec:results}
\begin{figure}[t]
	\begin{centering}
		\includegraphics[width=85mm]{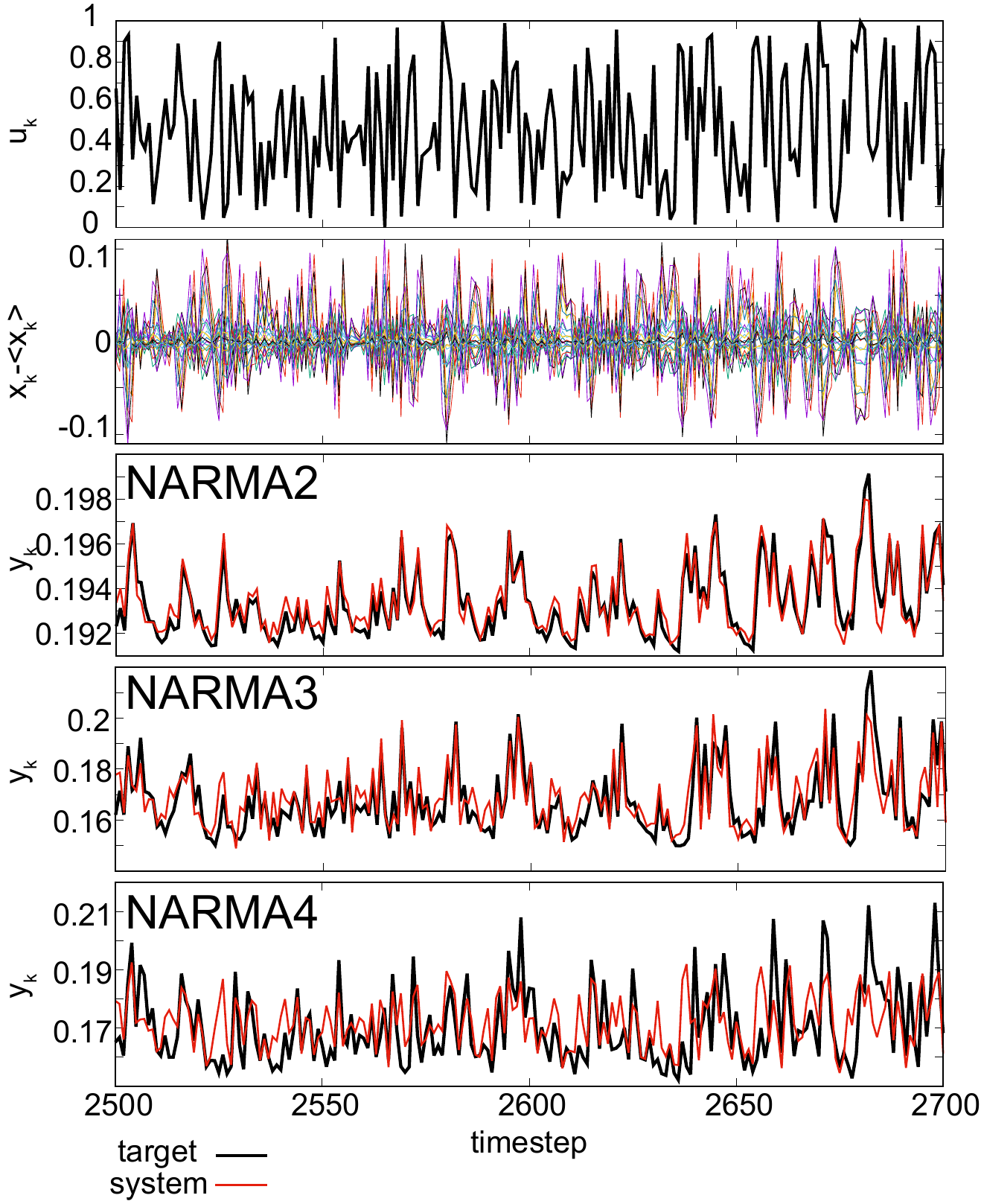}
		\par\end{centering}
	\caption{Typical performances for NARMA tasks. Random input sequence (the upper line), the corresponding reservoir dynamics from 90 computational nodes prepared from the arm (the second upper line), and the corresponding target and system outputs are plotted for each NARMA task. The case for $(A, \tau) = (6, 1)$ is shown for example.}
	\label{fig:NARMA_time_series}
\end{figure}

\begin{figure*}[t]
	\begin{centering}
		\includegraphics[width=180mm]{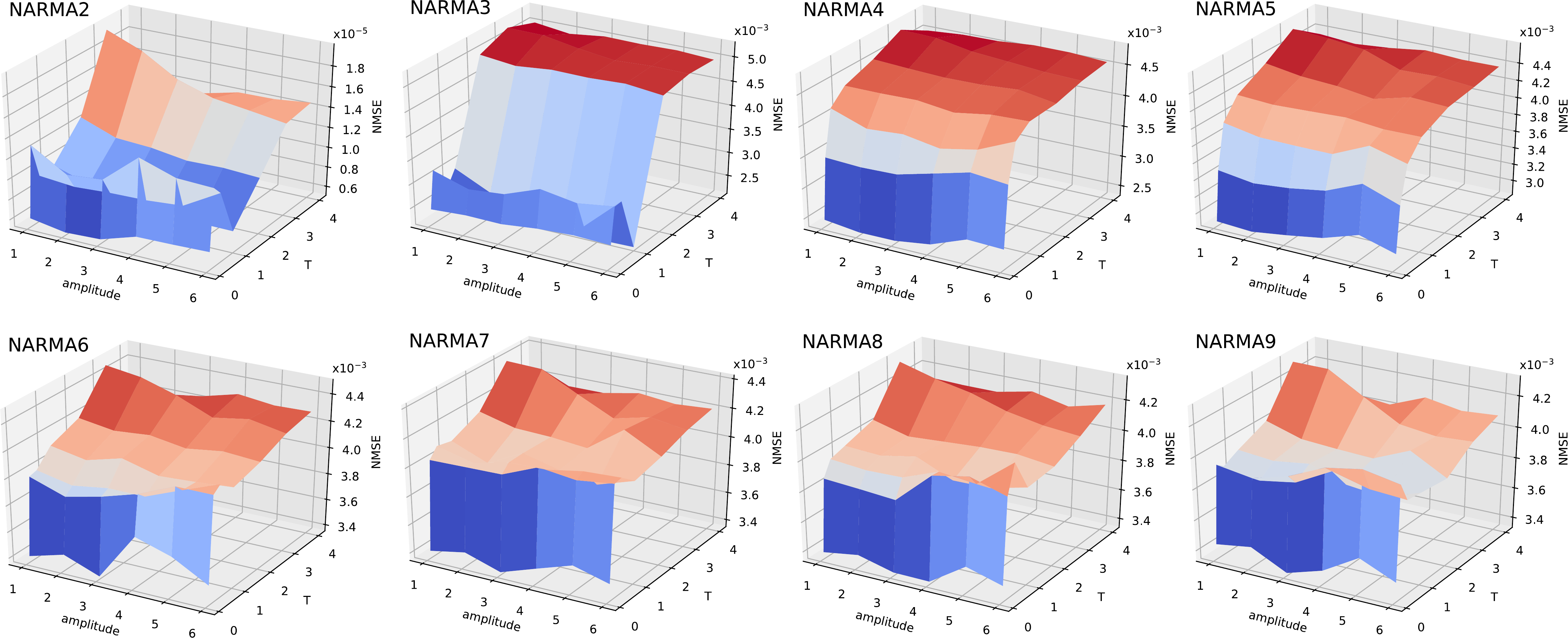}
		\par\end{centering}
	\caption{Analyses of the performance of NARMA tasks in terms of NMSE according to each setting of $(A, \tau)$.}
	\label{fig:NARMA_error}
\end{figure*}

In this section, the information processing capability of the soft continuum arm is investigated using numerical experiments. 
By using a benchmark task, which evaluates the capability to emulate nonlinear dynamical systems called {\it nonlinear auto-regressive moving average} (NARMA) system, and by assessing the linear and non-
linear memory capacity, we demonstrate how the parameter $(A, \tau)$ affects the performance of our system systematically, where $A$ is varied from 1 to 6 and $\tau$ is varied as 0.125, 0.25, 0.5, 1, 2, 3, and 4.
The evaluation schemes adopted here are popular in the context of recurrent neural network learning.

\subsection{Performance of NARMA tasks}
The NARMA task is a benchmark task that is commonly used to evaluate the computational capability of the learning system to implement nonlinear processing with long time dependence.
By calculating the deviations from the target trajectory in terms of errors, the NARMA task tests how well the target NARMA systems, which we introduce later, can be emulated by the learning system.
According to the choice of the target NARMA system, we can investigate which type of information processing can be performed in the learning system. 
The first NARMA system that we examine is a second-order nonlinear dynamical system \cite{NARMA}, expressed as follows:
\begin{equation}
	y_{k}=0.4y_{k-1} + 0.4 y_{k-1}y_{k-2} + 0.6u_{k}^{3}+0.1,
\end{equation}
and this system is called NARMA2 in this paper.
The next NARMA system is the $n$th-order nonlinear dynamical system, which is written as follows:
\begin{equation}
	y_{k}=ay_{k-1} + a' y_{k-1}(\sum_{j=0}^{n-1}y_{k-j-1}) + a''u_{k-n+1}u_{k}+a''',
\end{equation}
where $(a, a', a'', a''') = (0.3, 0.05, 1.5, 0.1)$.
The order of the system $n$ is varied from 3 to 9 in this experiment, and the corresponding systems are called NARMA$n$ systems for simplicity.
For the input stream to the NARMA systems, the range is linearly scaled from [-1, 1] to [0, 0.2] in order to set the range of $y_{k}$ into the stable range.
The performance is evaluated by comparing the system output with the target output in the evaluation phase using the {\it normalized mean squared error} (NMSE), expressed as follows: $NMSE= \frac{\sum_{k=2501}^{5000}(\hat{y}_{k}-y_{k})^{2}}{\sum_{k=2501}^{5000}\hat{y}_{k}^{2}}$, where $\hat{y}_{k}$ and $y_{k}$ are the target output and the system output at timestep $k$, respectively. 
For each setting of $(A, \tau)$, NMSEs for 20 trials, where the system is driven by a different random input sequence for each trial, are calculated and averaged for the analysis.

Figure~\ref{fig:NARMA_time_series} shows typical system outputs for the NARMA tasks in the evaluation phase, where the parameter is set to $(A, \tau) = (6, 1)$. 
We can see that, in particular for the NARMA2 emulation task, our system successfully traces the target NARMA system, and according to the increase of the order of the NARMA system, the overall task performance is gradually getting worse, suggesting the increase of the difficulty of the tasks.
In Fig.~\ref{fig:NARMA_error}, we have investigated the performance of the system in terms of NMSE systematically for each $(A, \tau)$ setting.
We can clearly observe that, according to the order of the NARMA system, the tendency of the performance differs by the selection of the parameter $(A, \tau)$.
For example, for the NARMA3 task, the case for $\tau = 1$ showed the lowest NMSE, which suggests the highest performance, while other tasks showed the highest performance when $\tau=0.125$. (The NARMA2 task also showed a good performance (although it was not the best) when $\tau=1$.)
These results imply that how to actuate the arm strongly affects the type of the information processing capability, which can be induced from the arm.
We should note that one common tendency observed was that the parameter $\tau$ affected more dominantly than the amplitude of the input $A$ to the task performance in this experimented parameter region.

\begin{figure}[t]
	\begin{centering}
		\includegraphics[width=85mm]{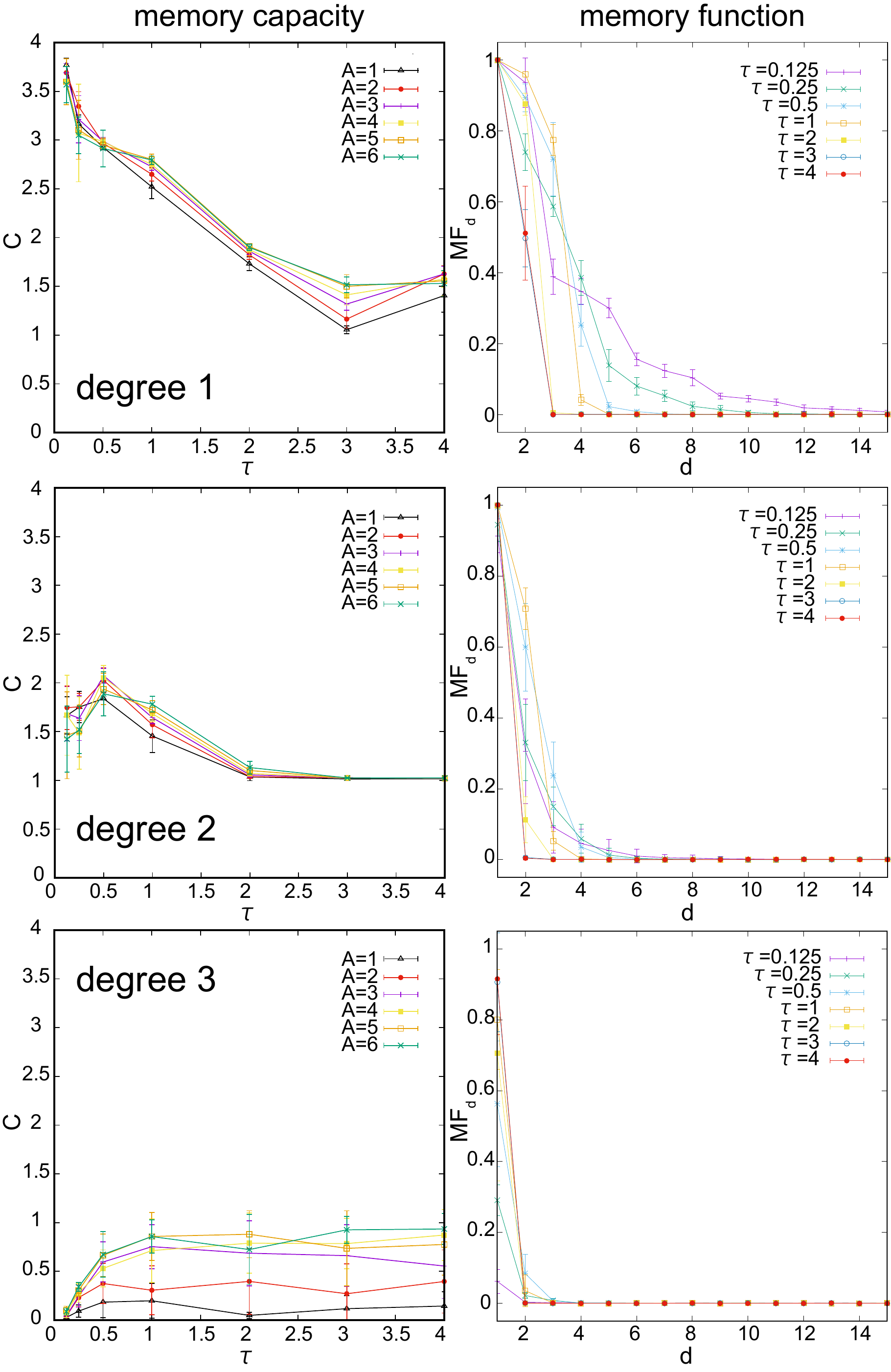}
		\par\end{centering}
	\caption{Analyses of linear and nonlinear memory capacities and typical memory function for each parameter setting. Error bars show standard deviation.}
	\label{fig:MC}
\end{figure}

\subsection{Analysis of Linear and Nonlinear Capacities}
Considering the fact that the information processing capability of reservoir dynamics can be characterized by its
property of transforming the input stream, it has been proposed to express the system's information processing capacity by evaluating its emulatability of nonlinear functions over the input stream \cite{IPC_2012}.
These functions are expressed as combinations of orthogonal functions, such as Legendre polynomials or Hermite polynomials, over each differently assigned delayed inputs. 
In this approach, the system's computational capability is decomposed into the degree of nonlinear processing and memory of the input that the system is capable to express. 
By exploiting the orthogonal functions, the computational capability can be safely decomposed into several nonlinear functions without overlaps. 
In this paper, we do not investigate all the combinations of Legendre polynomials of the previous inputs, but exploit $n$-th order Legendre polynomials for each delay of input expressed as:
\begin{equation}
	P_{n}(u_{k-d})=2^{n} \sum_{m=0}^{n} u_{k-d}^{m} \begin{pmatrix} n \\ m \end{pmatrix} \begin{pmatrix} \frac{n+m-1}{2} \\ n \end{pmatrix},
\end{equation}
where $\begin{pmatrix} n \\ m \end{pmatrix}$ is a binomial coefficient, and $u_{k-d}$ is the input of $d$ timesteps before from the timestep $k$.
In this study, we varied the value of $n$ from 1 to 10, and the delay $d$ is varied from 0 to 50 and investigated how our system is capable of learning each polynomial systematically.
Note that when $n = 1$, the polynomial becomes linear and it becomes equivalent to the case introduced in \cite{MC_2001}.

For the $n$-th order Legendre polynomials emulation task, using the system output time series, in each target function with given delay $d$, we calculate memory function of degree $n$, expressed as follows: $MF_{d}^{n} = \frac{cov^{2}(y_{k}, \hat{y}_{k})}{\sigma^{2}(y_{k})\sigma^{2}(\hat{y}_{k})}$, where $cov(x, y)$ and $\sigma(x)$ express the covariance between $x$ and $y$ and the standard deviation of $x$, respectively.
Then, the $n$-th order memory capacity $C_{n}$ can be expressed as follows: $C_{n} = \sum_{d=0}^{50} MF_{d}^{n}$.
By using the measure $MF_{d}^{n}$ and $C_{n}$, we aim to evaluate the information processing capability of the our system.
For each setting of $n$ and $d$ of the target Legendre polynomials, the learning scheme is exactly the same as explained previously, and, by using the 20 trials, the averaged $MF_{d}^{n}$ and $C_{n}$ are obtained for the analyses.

From the analyses of $C_{n}$, we first found that when $n$ is larger than 6, the value approaches to zero in all of the investigated parameter region.
This expresses the limitation of the expression power of the functions in our system.
Figure~\ref{fig:MC} shows the results for $n=1, 2,$ and $3$ for example.
As we saw in the NARMA task, according to the degree of nonlinearity $n$ of the target function, we can clearly observe that the parameter region showing the highest capacity differs.
For example, we can see the highest $C_{1}$ when $\tau=0.125$, while the highest $C_{2}$ can be observed in $\tau=0.5$ (Fig.~\ref{fig:MC}).
By increasing the degree of $n$ higher than 3, we tend observe the highest value in the parameter region where $\tau$ is larger than 1 and the input amplitude $A$ is larger than 4.
Basically, as we saw in the case for NARMA task, the behavior of $C_{n}$ was more dependent on the parameter $\tau$ than $A$, but for the degree of nonlinearity $n$ larger than 3, the effect of the input amplitude $A$ became dominant in the region of the high $\tau$.
Checking the $MF_{d}^{n}$ profile, we can see that, in this region, the memory function when $d=0$ is dominant, which means the immediate effect of input injection to the arm is exploited for the task performance (Fig.~\ref{fig:MC}). 

\section{CONCLUSIONS}\label{sec:conclusions}
Due to high compliance, soft continuum arms have immense potential for applications in spaces with human presence. However, as a result of high compliance (passive DoF), control of such manipulators is still an open challenge. Learning-based approaches such as morphological computation and reservoir computing have shown promise to surmount this problem. In this paper, we investigated the information processing capability of a PMA powered soft continuum arm. We applied randomly generated input signals of 5000 sample points with varying amplitude and bandwidth to an experimentally validated dynamic model and recorded the system response.  Using a benchmark task, which evaluates the capability to emulate nonlinear dynamical systems called {\it nonlinear auto-regressive moving average} (NARMA) system, and by assessing the linear and nonlinear memory capacity, we demonstrated how the parameter $(A, \tau)$ affects the performance of our system systematically. The results show that there are combinations of amplitude and bandwidth which operates the continuum arm in a domain most suited for reservoir computation based control implementation. In future work, we will extend these qualitative results to validate the findings experimentally.










\bibliographystyle{ieeetran}
\bibliography{refs}
\end{document}